\documentclass[runningheads]{llncs}
\usepackage{graphicx}
\usepackage{hyperref}
\usepackage{bm}
\usepackage[table]{xcolor}
\usepackage{colortbl}
\usepackage{booktabs,siunitx}
\usepackage{ulem}
\usepackage{amsmath}
\usepackage{csquotes}
\usepackage{siunitx}
\begin{document}
\title{Gaussian Process Emulators for Few-Shot Segmentation in Cardiac MRI}
%
%
\author{Bruno Viti\inst{1,5}\thanks{corresponding author: \email{bruno.viti@uni-graz.at}} \and Franz Thaler\inst{3,4} \and
Kathrin Lisa Kapper\inst{1,5} \and Martin Urschler\inst{2,5} \and
Martin Holler\inst{1,5} \and Elias Karabelas\inst{1,5}}
\authorrunning{B.Viti et al.}
%
\institute{Department of Mathematics and Scientific Computing, University of Graz, Graz, Austria\and Institute for Medical Informatics, Statistics and Documentation, Medical University of Graz, Graz, Austria \and Institute of Computer Graphics and Vision, Graz University of Technology, Graz, Austria \and Gottfried Schatz Research Center: Medical Physics and Biophysics, Medical University of Graz, Graz, Austria \and BioTechMed-Graz, Graz, Austria }
\maketitle              
\begin{abstract}
Segmentation of cardiac magnetic resonance images (MRI) is crucial for the analysis and assessment of cardiac function, helping to diagnose and treat various cardiovascular diseases. 
Most recent techniques rely on deep learning and usually require an extensive amount of labeled data. 
To overcome this problem, few-shot learning has the capability of reducing data dependency on labeled data. 
In this work, we introduce a new method that merges few-shot learning with a U-Net architecture and Gaussian Process Emulators (GPEs), enhancing data integration from a support set for improved performance. 
GPEs are trained to learn the relation between the support images and the corresponding masks in latent space, facilitating the segmentation of unseen query images given only a small labeled support set at inference. 
We test our model with the M\&Ms-2 public dataset to assess its ability to segment the heart in cardiac magnetic resonance imaging from different orientations, and compare it with state-of-the-art unsupervised and few-shot methods. 
Our architecture shows higher DICE coefficients compared to these methods, especially in the more challenging setups where the size of the support set is considerably small. The code is available on GitLab \footnote{\url{https://gitlab.com/bruno_viti/gpe_4_cardiac_fss}}.

\end{abstract}
\section{Introduction}
Medical imaging techniques like computed tomography (CT) and magnetic resonance imaging (MRI) are established technologies to assess patient health. 
Automatic medical image segmentation plays an important role by mapping anatomical structures to specific semantic labels which allows more in-depth analysis through follow-up applications~\cite{litjens2017survey}. 
This is especially relevant in cardiac imaging, with cardiovascular diseases being the world's leading cause of death \cite{World}. 
In recent years, deep learning (DL) models, and in particular Convolutional Neural Networks (CNNs), have been extensively employed to automate and accelerate heart segmentation, obtaining remarkable results in terms of accuracy~\cite{Chen2020}.
CNNs generally require a large amount of labeled data for training, posing a challenge in the medical field with a lack of data and limited manual segmentations. 
Moreover, CNNs assume that training and test data are drawn from the same distribution, i.e., that they are $i.i.d$. Therefore, in order for the model to perform well on a different data distribution, it would require a large-scale labeled dataset to update the networks' parameters for adaptation. Consequently, several DL approaches have been suggested to face the latter problem. 
For example, recent works proposed unsupervised domain adaptation or generalization architectures in the cardiac segmentation scenario~\cite{Gao2023,Kumari2023,Ouyang2022domaing,Chen2020improving}. 
These models alleviate the need for extensively labeled data in the target domain by leveraging knowledge from a related source domain where labeled data is more abundant. 
However, while these techniques can handle perturbations between training and test images from different modalities and sources, they are not designed to adapt to novel image orientation~\cite{Guan2021}. 
Another promising technique, which can effectively reduce data dependency, is few-shot segmentation (FSS)~\cite{Shaban2017,Chang2023}. 
In the FSS framework, the model is trained to learn how to segment a novel image, denoted as \emph{query}, having at its disposal only a small set of labeled examples, denoted as \emph{support}. 
Especially in the medical field, FSS is gaining attention, and several FSS architectures have been proposed~\cite{Hansen2022,Lin2023,Ouyang2022,Roy2020}. 
The method's effectiveness highly depends on the mechanism that extracts the information from the support set and integrates this new information with the query image. 
Most of the existing models rely on the prototype alignment paradigm, which tries to learn a common representation space, where the feature vectors of different object classes can be aligned, e.g.~\cite{Wang2019}. 
Differently, Johnander et al.~\cite{Johnander2022} proposed a novel approach based on GPEs, which they exploit to learn a mapping between dense local deep feature vectors and their corresponding mask values. 
Saha et al. \cite{Saha2022} also used FSS in combination with GPEs and applied their method to microscopy images. However, the majority of these approaches~\cite{Hansen2022,Lin2023,Ouyang2022,Roy2020,Wang2019} are limited to predicting binary segmentations. As we also show in this work exemplarily for \cite{Lin2023}, this results in a non-efficient class-by-class segmentation for multi-label query images~\cite{Hansen2023}. 
To face this issue, \cite{Hansen2023} introduced an extension of prototype alignment to perform one-step multi-class segmentation. 
They employ a self-supervised training approach, facing the challenge of limited labeled image data. 
However, in \cite{Hansen2023} they still consider the same type of images during training and testing, namely 2D short-axis cardiac MRI.\\
In contrast, our goal is to cope with the scarcity of data in the cardiac setting and to make segmentation models more adaptable to different cardiac image orientations. 
Therefore, we propose a model for multilabel cardiac image segmentation, which combines a U-Net-like architecture~\cite{Ronneberger2015} with GPEs as extractors of information from the support set~\cite{Johnander2022}. Specifically, we use the contracting branch of the U-Net to bring both the query and the support images into the latent space, in which the GPEs are trained to learn the relationship between the support images and the corresponding masks. 
This additional information is aggregated into the expanding branch, leading to a more accurate segmentation of the query.
We evaluate our model on the public M\&Ms2 dataset~\cite{Campello2021,Martin2023}, and assess its ability to generalize from 2D short-axis (SA) to long-axis (LA) cardiac MRI. 

\section{Methods}
\subsection{Problem Definition}
Few-shot segmentation aims to segment novel images given only a tiny set of labeled examples from the same distribution. 
In a fully supervised setting, training and test data are assumed to follow the same distribution, and further, each training instance is composed of an image and a segmentation mask.
In the FSS framework, each instance, denoted as \emph{episode}, includes a small set of labeled images, denoted as the \emph{support} set, and an unlabeled image, denoted as the \emph{query} set, both following the same distribution.
More formally, we denote the support set as $\{(\mathbf{X}_{\mathrm S, k},\mathbf{Y}_{\mathrm S,k})\}_{k=0}^{\mathrm K-1}$, including $\mathrm K$ pairs of images, $\mathbf{X}_{\mathrm S,k}\in \bbbr^{\mathrm H_0\times \mathrm W_0}$ and the corresponding segmentation masks, $\mathbf{Y}_{\mathrm S,k}\in \{0,1,\dots,\mathrm L-1\}^{\mathrm H_0\times \mathrm W_0}$ where $\mathrm L$ is the number of labels and $\mathrm H_0$,$\mathrm W_0$ refer to the height and width of the image, respectively. 
The task is to predict the segmentation mask $\mathbf{Y}_{\mathrm Q}$ of the query image $\mathbf{X}_{\mathrm Q}$.
During training, each episode includes a support and query set drawn from the domain distribution $\mathcal{D}_{\mathrm{tr}}$. 
For testing, the model predicts the segmentation mask of a novel image from the distribution $\mathcal{D}_{\mathrm{te}}$, having only a few examples of that distribution at its disposal.

\begin{figure}[t]
\includegraphics[width=\textwidth]{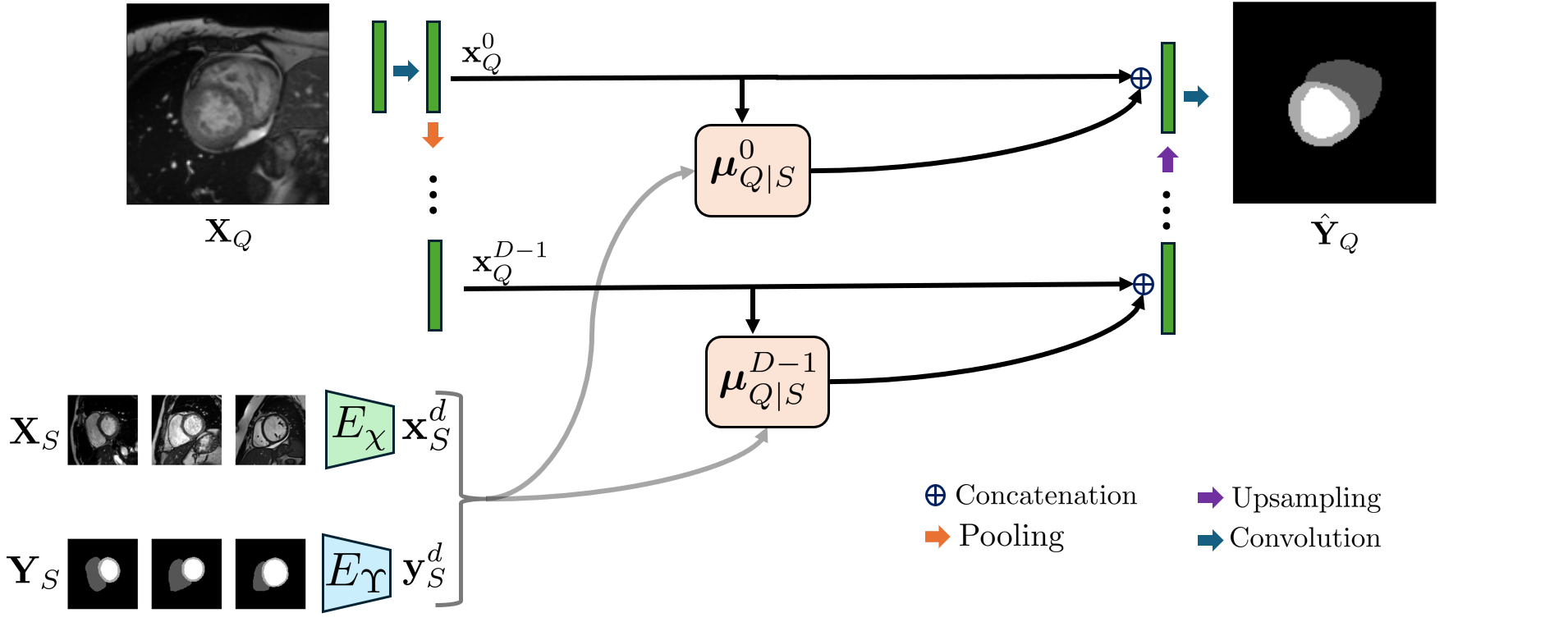}
\caption{Overview of a training episode of the proposed method. The encoder $E_{\chi}$ encodes the MRIs of the query and the support set, while $E_{\Upsilon}$ encodes the support masks. We train the GPE with the support's features, $\mathbf{x}_S^d$ and $\mathbf{y}_\mathrm{S}^d$. Then, given a new point $\mathbf{x}_\mathrm{Q}^d$, we infer the mean $\bm{\mu}_{\mathrm{Q}\vert \mathrm{S}}^d$. This additional information is then passed to the decoder $D_{\zeta}$ similar to a skip-connection, and the mask of the query input image is predicted. } \label{fig1}
\end{figure}

\subsection{Model Architecture}
Our network architecture is based on the work of \cite{Johnander2022}, and combines their GPE regressor with a U-Net architecture \cite{Ronneberger2015}. 
Although the U-Net has been chosen due to its effectiveness in handling medical images, the GPEs regressors are the fundamental components to extract the information from the support set and lead the model towards the correct segmentation of unseen images. 
Our model contains three major blocks, which are trained end-to-end, see Fig.~\ref{fig1}.

\subsubsection{Contracting path} 
As first step, we gradually extract and encode features from the input images while reducing their spatial dimensions. 
For each training instance, the input includes a query image $\mathbf{X}_{\mathrm Q}\in \bbbr^{\mathrm H_0\times \mathrm W_0}$, a support set of images $\mathbf{X}_{\mathrm S}\in \bbbr^{\mathrm H_0\times \mathrm W_0\times \mathrm K}$, and the corresponding masks $\mathbf{Y}_{\mathrm S}\in \{0,\dots,\mathrm L-1\}^{\mathrm H_0\times \mathrm W_0\times \mathrm K}$. 
The three inputs are moved into the latent space to extract their deep features. 
We employ the same feature extractor for $\mathbf{X}_{\mathrm Q}$ and $\mathbf{X}_{\mathrm S}$ as both are MRI. 
However, we need to train another encoder for the support masks $\mathbf{Y}_{\mathrm S}$. 
At the end of the contracting path, we have 
\begin{align*}
    \mathbf{x}_{\mathrm Q} &= E_{\chi}(\mathbf{X}_{\mathrm Q})\in \bbbr^{\mathrm H\times \mathrm W\times \mathrm F},\\
    \mathbf{x}_{\mathrm S} &= E_{\chi}(\mathbf{X}_{\mathrm S})\in \bbbr^{\mathrm H\times \mathrm W\times \mathrm K \mathrm F},\\
    \mathbf{y}_{\mathrm S} &= E_{\Upsilon}(\mathbf{Y}_{\mathrm S})\in \{0,\dots,\mathrm L-1\}^{\mathrm H\times \mathrm W\times \mathrm K \mathrm F},
\end{align*}
where $\mathrm F$ is the number of features.

\subsubsection{GPEs few-shot learning} 
GPEs offer a flexible framework for modeling complex functions and are suitable for working with small datasets~\cite{Rasmussen2006}. 
In our case, the training points are the support features, which can be reshaped as $\mathbf{x}_{\mathrm S} \in \bbbr^{\text{HWK} \times  \mathrm F}$, and ${\mathbf{y}_{\mathrm S} \in \{0,\dots,\mathrm L-1\}^{\text{HWK} \times \mathrm F}}$ is the corresponding output. 
With this reshaping, we assume that we have $\text{HWK}$ training points, and each of them lives in $\bbbr^{\mathrm F}$. 
As GPEs exhibit a computational complexity of $\mathcal{O}(N^3)$ for datasets of size $N$ \cite{Rasmussen2006}, our GPE will scale as $\mathcal{O}((\text{HWK})^3)$, making it necessary to work in a latent space in which $\mathrm H\ll \mathrm H_0$ and $\mathrm W\ll \mathrm W_0$. 
The GPE aims to predict $\mathbf{y}_{\mathrm Q}$ given a new input $\mathbf{x}_{\mathrm Q}$. 
The underlying assumption is that the two outputs $\mathbf{y}_{\mathrm S}$ and $\mathbf{y}_{\mathrm Q}$ are jointly Gaussian, i.e. 
\begin{align}
\left( \begin{array}{c}
\mathbf{y}_{\mathrm S} \\
\mathbf{y}_{\mathrm Q}
\end{array} \right) \sim \mathcal{N}\left(\left( \begin{array}{c}
\bm{\mu}_{\mathrm S} \\
\bm{\mu}_{\mathrm Q}
\end{array} \right), \left( \begin{array}{cc}
\mathbf{\Sigma}_{\text{SS}}&\mathbf{\Sigma}_{\text{SQ}}\\
\mathbf{\Sigma}_{\text{QS}}&\mathbf{\Sigma}_{\text{QQ}}
\end{array} \right)\right),
\label{eq1}
\end{align}
where $\bm{\mu}_{\mathrm S}$ and $\bm{\mu}_{\mathrm Q}$ are the prior means, which for simplicity are set to $\mathbf{0}$. 
The covariance matrices $\mathbf{\Sigma}$ encode the relationship between different points in the input space and quantify the similarity between them. 
Given two points $\mathbf{x}_i$,$\mathbf{x}_j\in \bbbr^{\mathrm F}$, the covariance matrix is defined by the kernel function $\mathbf{\Sigma}_{i,j} = k(\mathbf{x}_i,\mathbf{x}_j)$, where $k(\cdot,\cdot)\colon\bbbr^{\mathrm F} \times \bbbr^{\mathrm F} \rightarrow \bbbr$.
Ideally, for two similar points $\mathbf{x}_1$ and $\mathbf{x}_2$ we would expect similar outputs $\mathbf{y}_1$ and $\mathbf{y}_2$. 
In our model, we employ the squared exponential kernel function, due to its smoothness and interpretability, defined as
\begin{equation}
    k(\mathbf{x}_1,\textbf{x}_2)=s^2e^{-\frac{1}{2}\frac{||\textbf{x}_1-\textbf{x}_2||^2}{2l^2}},
\end{equation} 
where $s$ and $l$ are two additional hyperparameters that control the amplitude of the function values and its smoothness.
Using the Gaussian assumption in~\eqref{eq1}, we can infer the distribution of $\mathbf{y}_{\mathrm  Q}$ as
\begin{align*}
\mathbf{y}_{\mathrm Q}|\mathbf{x}_{\mathrm Q},\mathbf{x}_{\mathrm S},\mathbf{y}_{\mathrm S}\sim  \mathcal{N}(\bm{\mu}_{\mathrm Q| \mathrm S},\mathbf{\Sigma}_{\mathrm Q|\mathrm S}),
\end{align*}
where
\begin{align*}
    \bm{\mu}_{\mathrm Q| \mathrm S}&=\mathbf{\Sigma}^\top_{\text{SQ}}\left(\mathbf{\Sigma}_{\text{SS}}+\sigma^2\boldsymbol I\right)^{-1}\mathbf{y}_{\mathrm S},\\
    \mathbf{\Sigma}_{\mathrm Q|\mathrm S}&=\mathbf{\Sigma}_{\text{QQ}}-\mathbf{\Sigma}^\top_{\text{SQ}}\left(\mathbf{\Sigma}_{\text{SS}}+\sigma^2\boldsymbol I\right)^{-1}\mathbf{\Sigma}_{\text{SQ}}.
\end{align*}
Here, the term $\sigma^2 \boldsymbol I$ is used to bypass numerical issues and to account for noise in the data~\cite{Andrianakis2012}. 
The hyperparameter $\sigma$ represents the variance of noise that affects the observations. 
This hyperparameter, as well as $s$ and $l$, are learnable and therefore trained alongside the network weights. 
Thanks to the Gaussian assumption, we can infer the conditional mean of the output $\mathbf{y}_{\mathrm Q}$, which represents the closest prediction of $\mathbf{y}_{\mathrm Q}$ in the latent space according to the GPE~\cite{Johnander2022}. 

\subsubsection{Expanding path} 
We equip our model with an additional mechanism that is able to infer the posterior distribution of the query mask $\mathbf{y}_{\mathrm Q}$, given the image-mask pairs of the support set. 
At this point, we combine the deep features of the U-Net with the GPE prediction. 
To do that, we concatenate the posterior mean $\bm{\mu}_{\mathrm Q|\mathrm S}$ with the representation of the image in the latent space $\mathbf{x}_{\mathrm Q}$, obtaining $\mathbf{z}_{\mathrm Q} = (\bm{\mu}_{\mathrm Q|\mathrm S},\mathbf{x}_{\mathrm Q})$.
The latter is given as input to the decoder $D_{\zeta}$, which predicts the mask of the query image $\hat{\mathbf{Y}}_{\mathrm Q}=D_{\zeta}(\mathbf{z}_{\mathrm Q})$.
It should be noted that, as proposed in \cite{Johnander2022}, we can successfully leverage the skip-connection structure in the few-shot learning paradigm. 
In addition to learning the distribution of $\mathbf{y}_{\mathrm Q}$ in the deep latent space, we can do the same after each level of the U-Net encoder. 
More formally, given our encoders $E_{\chi}$ and $E_{\Upsilon}$ with depths $\mathrm D$, we denote by $\mathbf{x}_{\mathrm S}^{d}$ and $\mathbf{y}_{\mathrm S}^{d}$, the output of the encoders after $d$ levels. 
Following the previous terminology, we have $\mathbf{x}_{\mathrm S}^{\mathrm D-1}=\mathbf{x}_{\mathrm S},\ \mathbf{x}_{\mathrm S}^{0}=\mathbf{X}_{\mathrm S},\ \mathbf{y}_{\mathrm S}^{\mathrm D-1}=\mathbf{y}_{\mathrm S}$ and $\mathbf{y}_{\mathrm S}^0=\mathbf{Y}_{\mathrm S}$. 
The same idea applies to $\mathbf{x}_{\mathrm Q}$. 
Standard U-Net skip-connections concatenate the output of level $d$, $\mathbf{x}_{\mathrm Q}^d$, of the encoder, with the features in the decoder at the same level. 
We can combine this mechanism with our GPE regressor. 
For each level $d$ of the encoders, we assume that 
\begin{equation*}
    \mathbf{y}_{\mathrm Q}^{d}|\mathbf{x}_{\mathrm Q}^{d},\mathbf{x}_{\mathrm S}^{d},\mathbf{y}_{\mathrm S}^{d}\sim  \mathcal{N}(\bm{\mu}_{\mathrm Q|\mathrm S}^{d},\mathbf{\Sigma}_{\mathrm Q|\mathrm S}^{d}).
\end{equation*} 
At this point, we construct the corresponding $\mathbf{z}_{\mathrm Q}^d = (\bm{\mu}_{\mathrm Q|\mathrm S}^d,\mathbf{x}_{\mathrm Q}^d)$ and pass it to the level $d$ of the decoder $D_\zeta$. 
In the shallower levels, where $d<\mathrm D-1$, the resolution of the images and masks is too large; therefore, we down-sample them to the same size as used in the last level to make the GPEs computationally feasible.

\subsection{Training procedure}
The construction of meaningful query and support instances is crucial for FSS. 
In our setup, the model must be trained with different anatomical views to segment a novel view in the test set. 
To achieve that in the training phase, we treat SA slices from the base to the apex of the heart as different anatomical views based on their size, but with the same image orientation. 
Specifically, we divide our training dataset into $10$ new subsets based on the dimension of the heart, via counting the pixels of the ground truth. 
In this way, each subset contains cardiac MRIs with similar appearance and GPEs can be trained with different views. 
During training, each query-support instance is constructed with images from the same subset. 
For testing, we used a different orientation in the anatomical view, which corresponds to the LA slices.
\section{Experiments and Results}
We evaluate our method for a scenario where only SA 2D MRI slices are available with ground truth for training, while LA slices of different patients have to be predicted. 
In this scenario, it is beneficial to have a method that only requires few labeled LA slices in order to optimally segment the whole LA set. 
We denote our training set as $\{\mathbf{X}_{\mathrm{tr},i}\}_{i=0}^{\mathrm N-1}\in \mathcal{D}_{\mathrm{tr}}$ and our testing set as $\{\mathbf{X}_{\mathrm{te},i}\}_{i=0}^{\mathrm M-1}\in \mathcal{D}_{\mathrm{te}}$, where $\mathcal{D}_{\mathrm{tr}}=\{{\text{SA}}\}$, $\mathcal{D}_{\mathrm{te}}=\{{\text{LA}}\}$ hence $\mathcal{D}_{\mathrm{tr}}\ne \mathcal{D}_{\mathrm{te}}$. 
We compare our methodology with different baselines. 
We first measure the ability of a fully supervised method to generalize from SA to LA, having at its disposal only a few LA labeled samples. 
For this task, we use the widely used nnU-Net due to its good performance on a wide variety of medical image segmentation datasets \cite{Isensee2021}. 
Additionally, we compare our model to CAT-Net, a network for medical image FSS ~\cite{Lin2023}. 
Finally, we also utilize the unsupervised CSDG method \cite{Ouyang2022domaing}, a recently published method for domain generalization.

\subsubsection{Dataset}
For our evaluations, we use the M\&Ms-2 dataset~\cite{Campello2021,Martin2023}. 
The cohort contains 360 patients with different pathologies of the right ventricle and left ventricle, as well as healthy subjects. M\&Ms-2 is split into 200 patients for training and 160 for testing. 
Two pathologies are only present in the test set to evaluate the generalization to unseen pathologies. 
For each patient, the dataset provides 2 annotated LA MRI slices and about 20 annotated SA MRI slices. 
As described above, we use all of the SA slices for training and the LA slices for testing.

\subsubsection{Implementation Details}  
The two encoders, $E_{\chi}$ and $E_{\Upsilon}$,
contain five levels, each consisting of two convolutional blocks followed by $\mathrm{ReLU}$ activations, dropout, and batch normalization, ending with max pooling. 
For decoding, we use bilinear up-sampling at each level, followed by one convolution. 
The preprocessing involves intensity normalization
to zero mean and unit variance,  
and bilinear resampling to \qtyproduct{1.375 x 1.375}{\milli\meter} in-plane resolution. 
We use the ground truth mask to crop a centered bounding box with a fixed size of \numproduct{128 x 128} pixels. 
We train our model for \num{1500} epochs, adopting the Adam optimizer~\cite{kingma2017adammethodstochasticoptimization}.
We employ a weighted cross-entropy loss with weight \num{0.1} for the background and \num{0.3} for the rest of the labels. 
The batch size is set to \num{2}. 
We perform random data augmentation such as rotation and brightness adjustment. To train the network, we utilized an NVIDIA A100-SXM4-40GB. The training process took approximately \qty{24}{\hour} to complete.

\subsubsection{Comparisons with Other Methods}
We compare our method with three state-of-the-art methods, nnU-Net \cite{Isensee2021}, CAT-Net \cite{Lin2023}, and CSDG \cite{Ouyang2022domaing}. We trained CAT-Net using LA images. CAT-Net is trained through self-supervision with supervoxels, and at inference the true labels are given in the support set.  
CSDG is trained only with SA, being a fully unsupervised architecture, while nnU-Net is trained with a varying number of labeled LA slices added to the SA training data.
For nnU-Net, the number of shots in Tables~\ref{tab1}-\ref{tab2} refers to the number of LA images that are given during its training. 
These images are augmented to have a balanced training set between SA and LA.
\begin{table}[htbp]
    \caption{DICE (\%) scores for the 0/1-Shot setting.}\label{tab1}
    \small 
    \centering 
    
    \begin{tabular}{
        l
        *{5}{S[table-format=2.1, table-number-alignment=center,table-column-width=4em]}
      }
      \toprule
      Method & & 
      \multicolumn{4}{c}{DICE}   \\
      \cmidrule(r{1ex}){3-6}
      & {Shot}& {LV} & {RV} & {MY} & {Avg.}  \\
      \midrule
      \cellcolor{gray!30}CSDG\cite{Ouyang2022domaing} &0& 79.1 & 62.5 & 54.3 & 65.3 \\
      \cellcolor{gray!30}nnU-Net\cite{Isensee2021} &1& 43.8 & 30.6 & 36.5  & 40.0 \\
      \cellcolor{gray!30}CAT-Net\cite{Lin2023} &1& 71.7 & 58.8 & 32.5  & 54.2 \\
      \cellcolor{gray!30}Ours &1& $\mathbf{85.9}$ & $\mathbf{71.9}$ & $\mathbf{65.8}$ & $\mathbf{74.5}$ \\ 
      \bottomrule
    \end{tabular}
\end{table}
  \begin{table}[htp]
    \caption{DICE (\%) scores for different Shot settings.}\label{tab2}
    \setlength{\tabcolsep}{0pt}
    
    \begin{tabular*}{\textwidth}{
        @{\extracolsep{\fill}}
        l
        @{\hskip 1ex}
        *{3}{S[table-format=2.1, table-number-alignment=center]}
        @{\hskip 1ex}
        *{3}{S[table-format=2.1, table-number-alignment=center]}
        @{\hskip 1ex}
        *{3}{S[table-format=2.1, table-number-alignment=center]}
        @{}
        @{\hskip 1ex}
        *{3}{S[table-format=2.1, table-number-alignment=center]}
        @{}
      }
      \toprule
      \rowcolor{gray!30}
      \cellcolor{white!30}Method &
      \multicolumn{3}{c}{2-Shot} &
      \multicolumn{3}{c}{4-Shot} &
      \multicolumn{3}{c}{6-Shot} &
      \multicolumn{3}{c}{8-Shot} \\
      \cmidrule{2-4} \cmidrule{5-7} \cmidrule{8-10} \cmidrule{11-13}
      & {LV} & {RV} & {MY}   & {LV} & {RV} & {MY}   & {LV} & {RV} & {MY} & {LV} & {RV} & {MY}\\
      \midrule
      \cellcolor{gray!30}nnU-Net\cite{Isensee2021} & 64.1 & 47.6 & 53.2 & 79.0 & 66.1 & 68.7 & 81.8 & $\mathbf{76.1}$ & $\mathbf{73.0}$ & 85.7 & $\mathbf{76.9}$ & $\mathbf{74.6}$ \\
      \cellcolor{gray!30}Ours & $\mathbf{87.4}$ & $\mathbf{73.4}$ & $\mathbf{68.5}$ &$\mathbf{ 87.5}$ & $\mathbf{74.0}$ & $\mathbf{68.9}$ &$\mathbf{ 87.5}$ & 74.8 & 68.8 &$ \mathbf{87.6}$ & 75.3 & 68.9\\
      \bottomrule
    \end{tabular*}

  \end{table}The results in Table~\ref{tab1} show that nnU-Net fails to generalize from SA to LA having access to only one labeled image. 
In comparison, CSDG generalizes better to different MRI orientations, obtaining an average DICE of \qty{65.3}{\percent}.
Further, CAT-Net underperforms reaching only an average DICE of \qty{54.2}{\percent}.
This is likely because the LA images are 2D and thus, superpixel-based segmentations can only be generated in 2D which seems to be disadvantageous to the overall performance.
Differently to CAT-Net, our method efficiently exploits the information contained in the support image leading to an average score of \qty{74.5}{\percent} in the 1-Shot scenario, when relying only on a single LA image. 
In Table~\ref{tab2}, it can be observed that increasing the size of the support set up to 4, our method achieves the best result for all of the labels, outperforming nnU-Net on average by \qty{21.4}{\percent} in the 2-Shot scenario and by \qty{5.1}{\percent} in the 4-Shot scenario. 
When the size of the support set further increases to 6 and 8 images, our model continues to obtain the best score for the LV, while nnU-Net starts to perform better on the other two labels due to the additional LA images that are provided during training nnU-Net.
\begin{figure}
\includegraphics[width=\textwidth]{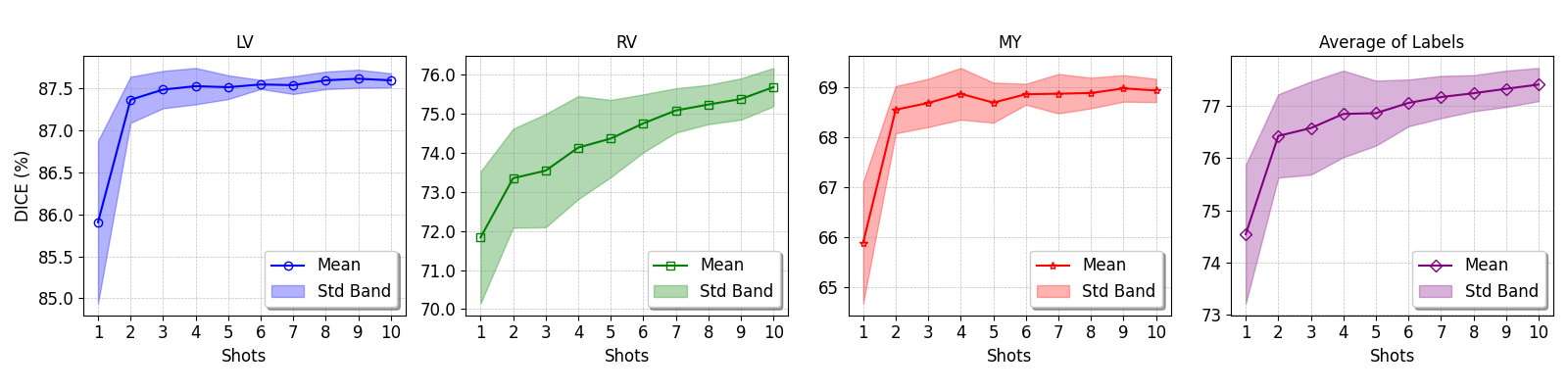}
\caption{Average DICE scores ($\%$) $\pm$ one standard deviation of our model for an increasing support set from 1 to 10 images.} \label{fig2}
\end{figure} 
Figures~\ref{fig2} and ~\ref{fig3} show that the performance of our model increasingly improves with the number of images in the support set.
The largest improvement for all labels is achieved when the size of the support set is increased from 1 to 2 images. 
While additional improvements beyond 2 support images are comparatively small for the LV and MY, the RV segmentation performance continues to benefit more, achieving an additional improvement of \qty{2}{\percent} in terms of DICE when increasing the support set from 2 to 10 images.
This behaviour is to be expected as the RV undergoes the largest variation from slice to slice and consequently, benefits the most from a diverse set of support images.\\
The impact of different support sets during inference is shown in Table \ref{tab3}, where the standard deviation of a 5-fold cross-validation for the support set is provided. 
As expected, in the 1-Shot scenario the model retrieves the LA information just from one image leading to a higher variability. 
Hence, in this scenario, the choice of images in the support set is significant. 
However, when the number of shots increases, the model is more robust to changes in the support set and the differences in the images affect the accuracy less.\\ 
\begin{table}[h]
    \caption{Standard Deviation (\%) scores for different Shot settings.}\label{tab3}
    \setlength{\tabcolsep}{0pt}
    
    \begin{tabular*}{\textwidth}{
        @{\extracolsep{\fill}}
        l
         @{\hskip 1ex}
        *{3}{S[table-format=0.1, table-number-alignment=center]}
        @{\hskip 1ex}
        *{3}{S[table-format=0.1, table-number-alignment=center]}
        @{\hskip 1ex}
        *{3}{S[table-format=0.1, table-number-alignment=center]}
        @{\hskip 1ex}
        *{3}{S[table-format=0.1, table-number-alignment=center]}
        @{\hskip 1ex}
        *{3}{S[table-format=0.1, table-number-alignment=center]}
      }
      \toprule
      \rowcolor{gray!30}
      \multicolumn{3}{c}{1-Shot} & 
      \multicolumn{3}{c}{2-Shot} &
      \multicolumn{3}{c}{4-Shot} &
      \multicolumn{3}{c}{6-Shot} &
      \multicolumn{3}{c}{8-Shot} \\
      \cmidrule{1-3} \cmidrule{4-6} \cmidrule{7-9} \cmidrule{10-12} \cmidrule{13-15}
       {LV} & {RV} & {MY} & {LV} & {RV} & {MY} & {LV} & {RV} & {MY}  & {LV} & {RV} & {MY} & {LV} & {RV} & {MY}\\
      \midrule
     1.0 & 1.7 & 1.2 & 0.3 & 1.3 & 0.5 &0.2 & 1.3 & 0.5 &0.1 & 0.7 & 0.2 &0.1 & 0.5 & 0.3\\
      \bottomrule
    \end{tabular*}
  \end{table}Finally, in Table \ref{tab4} the fraction of predicted images with a DICE score above a certain threshold is shown. In setting 1 we choose a threshold of $0.9$ for the LV and $0.7$ for the RV and MY. In setting 2 we fix the thresholds at $0.9,\ 0.75$ and $0.75$ for LV, RV and MY, respectively. While in setting 3 we choose as thresholds $0.9,\ 0.85$ and $0.75$.
\begin{table}[h]
 \caption{Fraction (\%) of predicted images with DICE above a certain threshold for different Shot settings.}\label{tab4}
    \small 
    \centering 
    
    \begin{tabular}{
        l
        *{4}{S[table-format=4.1, table-number-alignment=center,table-column-width=4em]}
      }
      \toprule
      Settings & 
      \multicolumn{4}{c}{Fraction \%}   \\
      \cmidrule{2-5}
      \rowcolor{gray!30}\cellcolor{white!30}& {1-Shot} & {4-Shot} & {6-Shot} & {10-Shot}  \\
      \midrule
      \cellcolor{gray!30}Setting 1 & 23.75 & 27.68 & 28.45 & 29.53 \\
      \cellcolor{gray!30}Setting 2 & 12.35 & 12.90 & 13.54 & 14.71 \\
      \cellcolor{gray!30}Setting 3 & 5.79 & 5.61 & 5.96 & 6.73 \\
      \bottomrule
    \end{tabular}
\end{table}
As expected, the number of predictions with meaningful segmentation increases with the number of shots, going from \qty{23.8}{\percent} to nearly \qty{30}{\percent} in the first scenario. 
\begin{figure}[b]
\includegraphics[width=\textwidth]{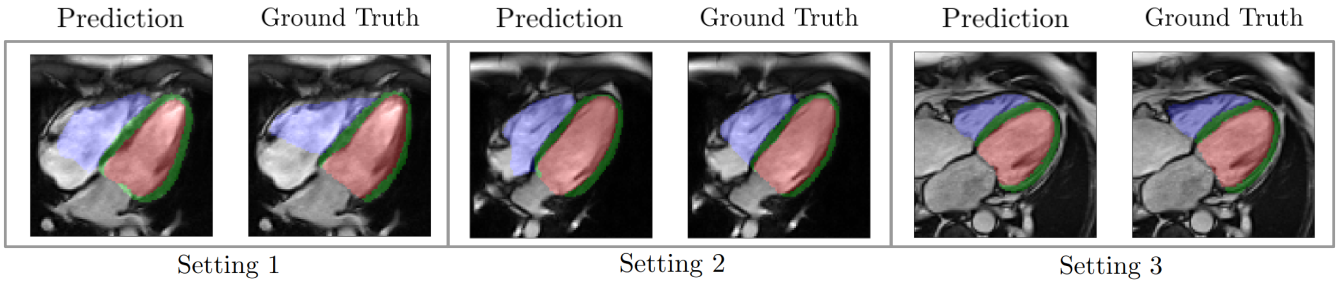}
\caption{Qualitative results of our method with different accuracies. On the left, we present a prediction that satisfies setting 1 but not settings 2 and 3, in comparison to the corresponding ground truth. In the center, a prediction meets setting 2 but not setting 3. On the right, the prediction fulfills setting 3.} \label{fig4}
\end{figure}
Figure \ref{fig4} depicts two caveats of the FSS method: the over-segmentation of the atrium, and the tendency of the myocardium to exhibit a ring-shaped pattern rather than an open-shaped configuration. The first problem is due to the absence of a clear gradient between the atrium and the ventricle. However, in most of the cases, the model is able to distinguish the two anatomical structures and segment only the ventricle. The second problem depends on the ring-shaped pattern of the myocardium in the short-axis images used for the training of the model, but, also in this case, extracting the information from the support set is beneficial in order to minimize this error. Moreover, we can argue that even if the percentage of predictions in the third scenario is relatively low, also the predictions in the other settings do not differ too much from the ground truth. However, the number of cases clinically usable is still small, and future works will aim to enhance the results.
\begin{figure}[h]
\includegraphics[width=\textwidth]{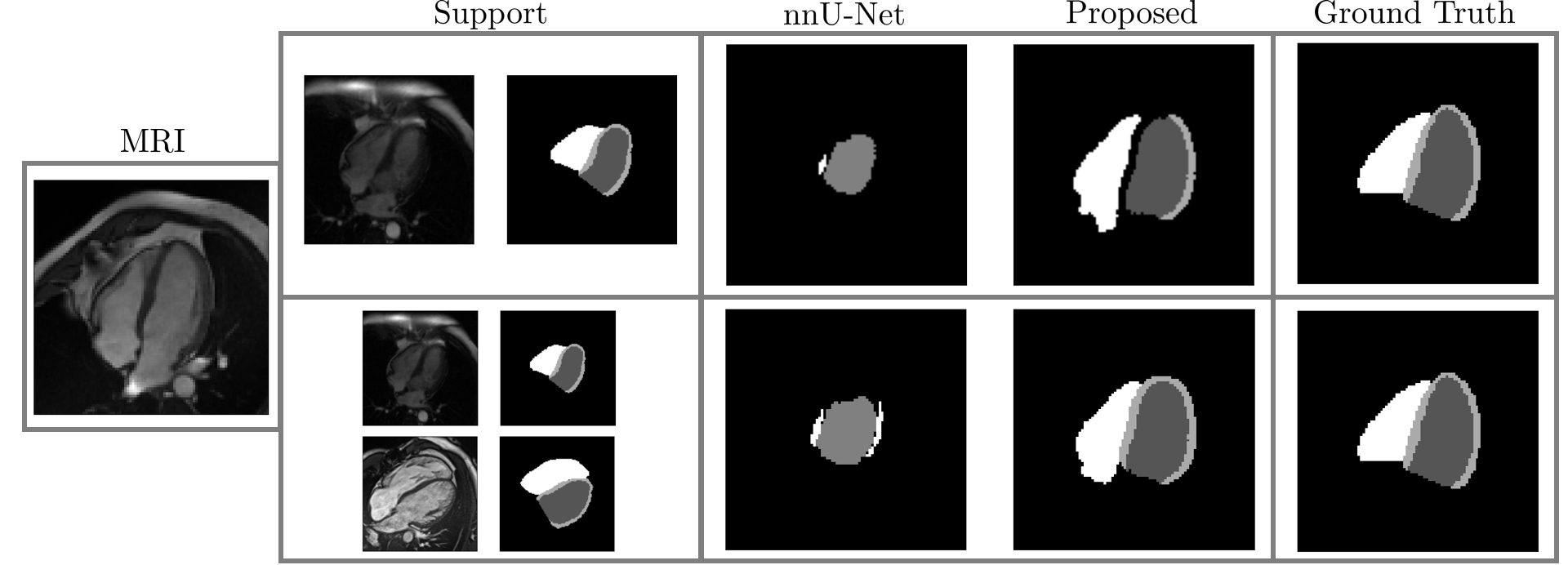}
\caption{Qualitative results of our method for an LA slice under 1-Shot (first row) and 2-Shot (second row). From left to right we show: MRI to segment, the support set used in the test phase, nnU-Net prediction, our prediction, ground truth.} \label{fig3}
\end{figure}
\section{Conclusions}
In this work, we proposed a few-shot cardiac segmentation architecture that combines the strengths of a U-Net with the flexibility of GPEs to segment different cardiac orientations with a limited amount of data. 
Our model incorporates information extracted from a limited support set with deep features of the CNN, increasing the generalizability of the whole architecture. 
We evaluated our method for a scenario in which only SA slices are available for training and obtained meaningful heart segmentations for LA orientation. 
In particular, we obtain a great performance boost using a single support image. As a result, with small effort in terms of annotation, it is possible to work with the differently oriented cardiac images without retraining of the whole network.
We plan to further improve our model, in particular, the GPE component by taking into account the conditional variance, and aim to explore the possibility of integrating uncertainty quantification in the predictions.
\section*{Acknowledgments and Disclosure of Funding}
B. V., K. L. K., and E. K. acknowledge funding from BioTechMed-Graz Young Research Group Grant CICLOPS; F. T. acknowledges funding from the Austrian Science Fund (FWF) CardioTwin grant I6540; M.U. acknowledges funding from FWF grant 10.55776/PAT1748423.

%
%
%
\bibliographystyle{splncs04}
\bibliography{references2}

\end{document}